%% file: elsarticle-template-1-num.tex
\documentclass[preprint, 12pt]{elsarticle}

\usepackage{color}

\newcommand{\inner}[1]{\left\langle#1\right\rangle}
\newcommand{\func}[1]{\text{#1}}

\usepackage{xspace}
\newcommand{\ie}{\textit{i}.\textit{e}.,\xspace}
\newcommand{\eg}{\textit{e}.\textit{g}.,\xspace}

\usepackage{graphicx}
\usepackage{amssymb}
\usepackage{amsthm}
\usepackage{amsmath}
\usepackage{enumitem}
\usepackage{lineno}
\usepackage{hyperref}
\usepackage{caption}
\usepackage{subcaption}
\usepackage{rotating}

\renewcommand\vec{\mathbf}
\usepackage{listings}

\journal{Computational Toxicology}

\begin{document}

\begin{frontmatter}


\title{Understanding Adverse Biological Effect Predictions Using Knowledge Graphs}



\author[A,B,C]{Erik B. Myklebust}
\author[D,C]{{Ernesto} {Jim\'{e}nez-Ruiz}}
\author[E]{{Jiaoyan} {Chen}}
\author[A,F]{{\\Raoul} {Wolf}}
\author[A,G]{{Knut Erik} {Tollefsen}}

\address[A]{Norwegian Institute for Water Research (NIVA), Oslo, Norway}
\address[B]{NORSAR, Kjeller, Norway}
\address[C]{SIRIUS, University of Oslo, Oslo, Norway}
\address[D]{City, University of London, London, United Kingdom}
\address[E]{University of Oxford, Oxford, United Kingdom}
\address[F]{Norwegian Geotechnical Institute (NGI), Oslo, Norway}
\address[G]{Norwegian University of Life Sciences (NMBU), \AA s, Norway}

\begin{abstract}
Extrapolation of adverse biological (toxic) effects of chemicals is an important contribution to expand available hazard data in (eco)toxicology without the use of animals in laboratory experiments. In this work, we extrapolate effects based on 
a knowledge graph (KG) consisting of the most relevant effect data as domain-specific background knowledge. An effect prediction model, with and without background knowledge, was used to predict mean adverse biological effect concentration of chemicals as a prototypical type of stressors. The background knowledge improves the model prediction performance by up to 40\% in terms of $R^2$ (\ie coefficient of determination). We use the KG and KG embeddings to provide quantitative and qualitative insights into the predictions. These insights are expected to improve the confidence in effect prediction. 
Larger scale implementation of such extrapolation models should be expected to support hazard and risk assessment, by simplifying and reducing testing needs. 

\end{abstract}


\end{frontmatter}

\section{Introduction}

Extrapolation {using} already available adverse biological (toxic) effects to species and stressors, which are not readily available (\eg new chemicals and/or species) or {are} ethically questionable (\eg  vulnerable or endangered species), is an important development towards more sustainable testing approaches in (eco)toxicology. 
{According to \citet{animaluse, moreanimaluse}}, 1 million animals are used in laboratory experiments each year in the United States alone. Moreover, \citet{Busquet_Kleensang_Rovida_Herrmann_Leist_Hartung_2020} puts this number between 10 and 18 million in the European Union. 

We introduce a novel approach for extrapolation using a stressor-effect informed knowledge graph (KG) and its vector embeddings to provide additional information to the extrapolation models.

{To this end}, we have created the Toxicological Effect and Risk Assessment KG (TERA), which integrates key data concerning chemicals and species (\eg~{categorization})
used in laboratory experiments as well as the experiments themselves \citep{Myklebust2019KnowledgeGE,swj_paper}.

Knowledge graph embedding (KGE) models {such as DistMult and TransE} \citep{KGE_survey_2017, DBLP:journals/tkdd/RossiBFMM21} can be applied to TERA to extract low-dimensional vector representations of its entities. These
vector representations enable the use of standard machine learning prediction models to extrapolate effect data from existing chemical-species pairs to new combinations \citep{Myklebust2019KnowledgeGE,swj_paper}.

TERA can also help us understand the prediction based both on the vectors and the structure of the graph itself, \eg neighbourhoods. These explanations are very important when using models in downstream tasks (\eg ecological risk assessment) to increase the confidence in predictions. 

This paper builds on our previous work, in \citet{Myklebust2019KnowledgeGE,swj_paper}, by moving from a strictly
(binary) classifications of effects to a prediction of chemical concentrations
for a given effect in a specific organism, \ie which concentration is toxic (lethal) to an organism. 
In this paper we restrict the data coverage to organic compounds, in order to yield a more consistent data set. We
have also built on the KGE analysis section in \citet{swj_paper} to provide
insight and interpretations for the predictions. This 
works contributes to a 
larger effort of making effect prediction models explainable.
%
The work presents the results of gap-filling tasks, one for chemicals and one for species. We perform chemical-species pair predictions to fill (synthetically) missing adverse effect data. 
We show that using knowledge graph embeddings in this gap-filling prediction task can improve both the results and the interpretation of a complex model.

The rest of the paper is organized as follows. Section \ref{sec:pap3_background} provides relevant background information on effect extrapolation and introduces the concepts {of KG, KGE}, and Explainable AI (XAI). Section \ref{sec:pap3_method} provides {an} overview of the methods while the specific details are found in \ref{sec:method_appendix}. Section \ref{sec:pap3_results} introduces the metrics used to evaluate the models followed by the evaluation and the explanations of {the predictions of these models}. Section \ref{sec:pap3_discussion} discusses the implications of the results, while Section \ref{sec:pap3_conclusion} concludes the work and indicates {some} future research directions. 

\section{Background and State-of-the-Art}
\label{sec:pap3_background}

\subsection{Effect {E}xtrapolation}
In ecological risk assessment, it is not feasible to conduct experiments on all possible species-chemical combinations ($S \times C$), therefore, extrapolation models have been developed, such as Quantitative {S}tructure-Activity {R}elationship (QSAR) \citep{BRADBURY1995229}, and read-across\footnote{Use data from one sub-domain as predictor for another.} methods {have} been designed, {such as} \citep{https://doi.org/10.1002/qua.26097} that {uses} predicted features for chemical hydrophobicity in a linear regression model. 
Most of these models consider one (or a small group treated as one) species and extrapolates on relatively small groups of chemicals. This is to remove inter-species variability, however, 
intra- and inter-assay variability have been shown to be large \citep{10.1007/978-3-642-46856-8_35}.

This work goes under the (whole organism) read-across methodology where we try to mitigate {its} shortcomings. Standard read-across is done by creating similarity measures (unique to the method), \eg chemical fingerprint similarity, among species or chemical to justify the read-across (\eg \citet{pmid28172359}). 
Read-across is more difficult in the species domain than in the chemical one due to the lack of defining features such as molecular weight or partition coefficient, although, genetic similarity has emerged as a viable option. One of these tools is the Sequence Alignment to Predict Across Species Susceptibility (SeqAPASS; \citet{seqapass,10.1093/toxsci/kfy186}). This tool uses available data on phylogenetic conservation of proteins across different taxa to indicate the likelihood of the existence of similar biological targets in other species. It only provides a sequence similarity of a given molecular target (\eg a protein) or the ligand-binding or ligand-protein interacting site, and thus does not directly infer the toxic potency of the chemical for the whole organism.




\subsection{Semantic Web technologies and Knowledge Graphs}

The main purpose of using {Semantic Web technologies} in this work is that it  facilitates the integration {and access of disparate sources}, \eg tabular and graph data. These disparate resources are integrated within the TERA knowledge graph~\citep{swj_paper}.


We follow the RDF-based notion of knowledge graphs \citep{j.websem510} which are composed by 
relational facts/triples 
$\left\langle s, p, o \right\rangle$,
where $s$ represents a subject (a class or an instance, \eg \texttt{Bluegill}), 
$p$ represents a predicate (a property, \eg \texttt{eol:Habitat}) 
and $o$ represents an object
(a class, \texttt{freshwater biome}, an instance or a data value).
RDF entities (\ie classes, properties and instances) are represented by an URI (Uniform Resource Identifier).

RDF-based KGs like TERA enable the use of the available Semantic Web infrastructure, including reasoning and the SPARQL query language. 
SPARQL uses graph patterns as a query, which is a large benefit for hierarchy-centric data such as taxonomies. RDF and SPARQL are (Semantic) Web standards defined by the W3C.\footnote{\url{https://www.w3.org/standards/semanticweb/}}

\subsection{KG Embeddings}

{A KG is a symbolic representation and could not be easily or directly used in machine learning which requires sub-symbolic data representation.}
Therefore, models for creating KG embeddings have emerged. An embedding is a low dimensional representation of {an element in the KG, with its semantics kept in the vector space.} 

Knowledge graph embedding methods aim at reducing the complexity of a KG into a smaller (vector) space. This can be done in two main ways. The first is based on walking the KG in some fashion to create sentences and using established {word embedding} models, \eg RDF2Vec \citep{DBLP:journals/semweb/RistoskiRNLP19} and OWL2Vec* \citep{chen2020owl2vec} use {Word2Vec} \citep{word2vec} on the generated sentences. 
The other {methods model the truth of a triple (\ie a fact) in the KG by a probability}. In this work, we use the latter, such that the aim of a model is to find a (scoring) function $S(t)\mapsto \mathbb{R}$ where $S$ is proportional to the probability that a triple $t$ is true. An example of such a model is TransE \citep{NIPS2013_5071}, which represents the predicate in a triple as a transformation from the subject to the object of a triple. Formally, the model minimizes {the overall loss of a given set of triples (or all the existing triples) of the KG, where each triple contributes to a loss of}
$||\vec{s} + \vec{p} - \vec{o}||_2$,
$\vec{s},\vec{p},\vec{o} \in \mathbb{R}^k$ {denote the vector representations of the subject ($s$), predicate ($p$) and object ($o$) of a triple ($\left\langle s, p, o \right\rangle$), respectively. $k$ is the embedding dimension.}

\subsection{{Explainable AI}}
Explainable AI (XAI) has recently {attracted wide attention in many research communities} (\eg \citet{vilone2020explainable,DBLP:journals/inffus/ViloneL21,DBLP:series/ssw/47,chen2018knowledge}). 
Models traditionally used in effect prediction are inherently explainable as they are mostly based on linear regression or {decision tree-based} approaches. However, deep learning is encroaching on this and there is a larger push in the community for the use of data-centric approaches, {due to its better performance} \citep{WITTWEHR2019100114}.

Certain prediction models can be black boxes but by using a knowledge graph we can provide an explanation to the user of the model \citep{info11020122}. These explanations can be presented to {experts who} can verify if a model prediction is consistent with their expectations. 


In this work, we consider steps toward explanations using the knowledge graph structure both quantitatively and qualitatively, which are explained in the next section.  

\section{Method}
\label{sec:pap3_method}
This section introduces the knowledge graph and embeddings along with the prediction model and methods for gaining insight. Appendix \ref{sec:method_appendix} provides details. 

\subsection{The TERA Knowledge Graph} \label{sec:pap3_tera}
In \citet{Myklebust2019KnowledgeGE} we introduced the Toxicological Effect and Risk Assessment Knowledge Graph (TERA) and it consists of three distinct parts{:}
\begin{enumerate}
    \item Effects sub-KG. This sub-KG contains {knowledge from the ECOTOXicology Knowledgebase (ECOTOX KB) \citep{ecotox}, where the original tabular datasets are transformed into triples.}
    \item Chemical sub-KG. This part comprises chemically related data, {which} includes selected data from ChEMBL \citep{10.1093/bioinformatics/btt765}, Chemical Entities of Biological Interest (ChEBI) \citep{chebi}, Medical Subject Headings (MeSH) \citep{mesh}, and PubChem \citep{pubchem}. 
    \item Species sub-KG. {This part includes knowledge transformed from two distinct taxonomies: the National Center for Biotechnology Information (NCBI) Taxonomy and the Encyclopedia of Life (EOL) traits \citep{eol}.}
\end{enumerate}

The sources mentioned above are disparate and {need} to be aligned. We used ontology alignment systems {such as LogMap} \citep{logma_ecai2012,logmap2011} and external sources {such as} Wikidata \citep{wikidata2014}, to construct {the} alignments among sources.
The integration of new sources in to TERA can vary in effort from simple, where exact mappings exist (\eg using Wikidata), to a highly manual process if there are mismatches between sources or multi-level merging is needed (\eg ECOTOX species to NCBI Taxonomy).

\subsection{Knowledge {G}raph {E}mbedding {Model}}

Based on {our} previous work in \citet{swj_paper}, we found that the model ComplEx \citep{DBLP:journals/corr/TrouillonWRGB16} is best suited to embed TERA and, therefore, we only use this model in this work. ComplEx performed consistently well over all data sampling strategies.
ComplEx models the probability of a fact as the inner product of complex vectors $\vec{e}_s,\vec{e}_p,\vec{e}_o \in \mathbb{C}^k$, \ie
\begin{align}
    \begin{split}
    S_{\text{ComplEx}}(s,p,o) &= \inner{\vec{e}_s,\vec{e}_p,\vec{e}_o}\\ 
    &= \inner{\Re(\vec{e}_s)+i\Im(\vec{e}_s),\Re(\vec{e}_s)}\\
    &+ \inner{i\Im(\vec{e}_s),\Re(\vec{e}_p)+i\Im(\vec{e}_o)} \\
    &= \inner{\Re(\vec{e}_s),\Re(\vec{e}_p),\Re(\vec{e}_o)} \\
    &+ \inner{\Im(\vec{e}_s),\Re(\vec{e}_p),\Im(\vec{e}_o)} \\
    &+ \inner{\Re(\vec{e}_s),\Im(\vec{e}_p),\Im(\vec{e}_o)} \\
    &+ \inner{\Im(\vec{e}_s),\Im(\vec{e}_p),\Re(\vec{e}_o)} 
    \end{split}
\end{align}
where $i=\sqrt{-1}$ and, $\Re(x)$ and $\Im(x)$ are the real and complex parts of $x$, respectively. The optimization of this model is described in \ref{sec:complex_appendix}.

\subsection{Adverse Effect {D}ata}
We focus on {the} acute mortality {of} test organisms in this work. We use the ECOTOX KB which is the largest publicly available collection of ecotoxicological effect data. This data is integrated into TERA ({see} details in Section \ref{sec:pap3_tera}) and can be accessed via SPARQL queries. 
In short, we gather results with endpoint (experimental steady state) in $\{LC_{50},LD_{50},EC_{50}\}$\footnote{Only $EC_{50}$ where effect is mortality.} and experimental duration of $\geq 24h\ \wedge \leq 96h$. See Query~\ref{lst:query} in \ref{sec:query_appendix}. Since the effects are acute{,} the relevance of experimental duration is assumed to be mitigated. 
To limit noise in the data, we only consider compound-species pairs with three or more results, and we use the median (of results for each compound-species pair) of these as the ground truth.\footnote{There is not \emph{one} truth in this case, but this is a close approximation.}

\subsection{Grouping {C}hemicals and {S}pecies}
\label{sec:grouping}
{We group the species and chemicals that are to be used in} the gap-filling task. This is to test the {model's ability of transferring} knowledge from one group to another. 
{The species are grouped according to the divisions in the taxonomy} In this work, we consider the {following species} 
groups as defined in ECOTOX: Fish,
       Crustaceans,
       Insects/Spiders,
       Amphibians,
       Worms,
       Invertebrates, and
       Molluscs.
       
The chemical hierarchy is {more complicated} and we need to choose a different method. {For example,} DEET (\emph{N,N-Diethyl-m-toluamide}, CAS 134-62-3) is both a pesticide and an organic compound. 
We ({hierarchically}) cluster {the} chemicals based on the PubChem fingerprints \citep{pubchem}. These fingerprints are defined using $881$ binary features of a compound. 

\begin{figure}
    \centering
    \includegraphics[width=0.6\textwidth]{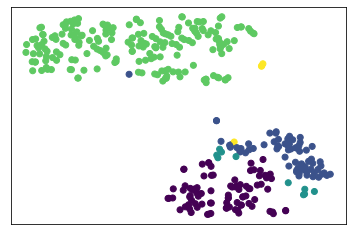}
    \caption{2D representation of organic chemical clusters. Note that, scale of this figure is irrelevant.}
   \label{fig:organic_clusters_fp_sim}
\end{figure}

We fix the number of clusters {to} 5. This was chosen to retain a certain size of the smallest cluster (yellow in Figure \ref{fig:organic_clusters_fp_sim}). 
The clusters can be seen in Figure \ref{fig:organic_clusters_fp_sim}. Note that, hierarchical clustering is performed in the original space ($\{0,1\}^{881}$) but have been reduced to $\mathbb{R}^2$ in the figure using t-SNE \citep{JMLR:v9:vandermaaten08a} and, therefore, {there is some overlap between clusters in the figure}. 

\subsection{Prediction}

The prediction model used in this work is Support Vector Machine (SVM) \citep{Scholkopf99kernelprincipal} with a Radial basis function (RBF) kernel (additional details provided in \ref{sec:svm_appendix}). This {is} chosen as it is robust to large amounts of noise (uncertainty) as is the case in this work. 
We use the \texttt{scikit-learn} package \citep{scikit-learn} and tune the most important model parameters ($C$ and $\gamma$ will influence regularization and kernel scaling, respectably) by performing a grid search with $C \in \{10^k,k\in\{-4,4\}\}$ and $\gamma \in \{10^k,k\in\{-4,4\}\}$.

The input to the model is a vector resulting from the concatenation of the vectors representing the chemical and the species.\footnote{Experimental features, such as experimental duration, organism details, experimental setup etc. can be added, however, here we focus on only the embeddings} The vector representations are either generated by the KGE models or a random projection into a vector space of the same size (as defined by in the KGE, we use $400$ in this work). Note that, we train four different initializations of ComplEx with the best hyperparameters found in~\citet{swj_paper} and with vectors of size 100; then these vectors are concatenated to obtain vectors of size 400 for each chemical and species entity.  
This limits the effect that comes from the random model initialization as this has been shown to have large influence on KGE models \citep{xu2021multiple}. 
The output of the prediction model is $-log_{10}(y)$, where $y$ is the chemical concentration.

\subsection{Toward KG-driven Insights}

Providing insights into predictions is very important in the verification of model validity. We explore two different ways of generating {prediction insights}:

\begin{enumerate}
    \item Neighbourhood density. \ie how many graph nodes are within a certain (Euclidean) radius of the prediction species and chemical. We look at both entity density\footnote{The number of entities in the neighbourhood.} and data density.\footnote{The number of (experimental) data points in the neighbourhood. This is relevant as not all entities have associated data.}
    \item Error prediction. Certain properties of the graph, including neighbourhood can be a potential predictor of the prediction error. 
    We predict the error based on the (Euclidean) distance (similarity) among chemicals and species using a Random Forest model. 
    \item Common facts with close entities. {We find} the closest entities to the prediction species and chemical based on {their embeddings using euclidean distance (or cosine similarity) and then find} common facts among them. This will give an indication of whether relevant or generic facts influence the prediction. 
\end{enumerate}

The first two are analogous and provide confidence in the prediction, while the last will give more insights into which facts affect the prediction. Algorithms describing the three methods are found in \ref{sec:algo_appendix}.

\subsection{Workflow}
\label{sec:pap3_workflow}

\begin{figure}
    \centering
    \includegraphics[width=0.95\textwidth]{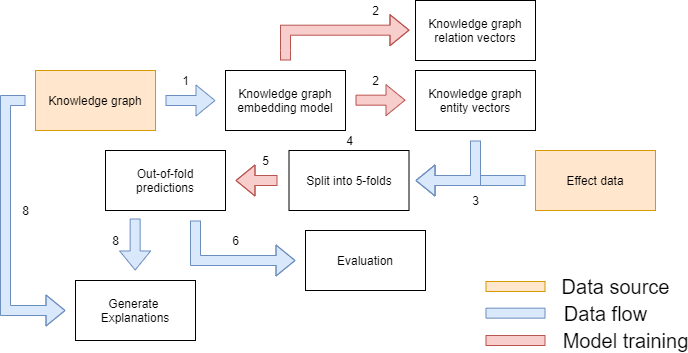}
    \caption{Flowchart of steps taken to produce results. The numbers correspond to the steps taken in Section \ref{sec:pap3_workflow}.}
    \label{fig:flowchart}
\end{figure}

In Figure \ref{fig:flowchart} we show the steps taken to produce the results and explanations for the chemical effect prediction. These can be summarized as follows:
\begin{enumerate}
    \item TERA is feed into the KGE model which samples positive and negative triples to use for training. 
    \item After training, the KGE has produced vector representations of the entities and relations in TERA.
    \item Using the biological effect data, we extract the vectors for compound and species relevant for each sample. 
    \item \label{th:five_fold} The samples are divided into five subgroups/folds (equal parts) where there are no overlap in the relevant group (\ie chemical group or species division as described in \ref{sec:grouping}.). 
    \item The SVM is trained by leaving one fold out and training on the other four. This is repeated for all five folds. 
    \item \label{th:evaluate} This process produces so-called out-of-fold predictions which are used to evaluate the model. We produce $R^2$ and categorical accuracy as described in Section \ref{sec:pap3_metrics}. 
    \item Steps \ref{th:five_fold} to \ref{th:evaluate} are repeated 100 times with random folds to produce mean and standard deviation. 
    \item Explanations are produced using the average predictions from the 100 randomly initialized runs. 
\end{enumerate}

\section{Results}
\label{sec:pap3_results}

We aim at evaluating the usefulness of using knowledge graph (embeddings) for effect prediction. We evaluate the problem of gap-filling, where sub-sets of chemicals or species are unknown and need to be predicted.  

\subsection{Challenges}
The results presented in this study heavily relies on the quality of effect data and the completeness of TERA. However, as shown in Figure \ref{fig:std_input}, the standard deviation of experiments can be rather large. In addition, we know that currently TERA does not contain all knowledge related to the domain of effect prediction. This is a limiting factor, however,
increasing the coverage of the domain without a proper validation
might reduce performance by 
including
non-relevant,
inconsistent, or noisy information. 

\subsection{Data sampling strategies}

\begin{figure}
    \centering
    \includegraphics[width=0.6\textwidth]{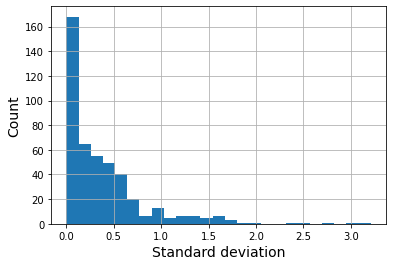}
    \caption{Distribution of standard deviation (log-concentration) of each organic chemical-species pair.}
   \label{fig:std_input}
\end{figure}

The experimental data in ECOTOX contains ambiguous samples.
We demonstrate this by calculating the standard deviation (of log concentration) of the experimental results for each chemical-species pair (with $\geq 3$ results). The distribution of this is shown in Figure \ref{fig:std_input}. We see that the majority of results are within $0.5$ orders of magnitude. However, up to 10\% have a standard deviation larger than one order of magnitude. These large discrepancies will make some predictions harder than others. 

The data is sampled as a gap-filling problem, \ie we remove a proportion of each group from the training data for testing purposes.

The next subsections presents the results for gap-filling species and chemicals results. The figures present average metrics (next section) and standard deviation over 100 random permutations of the unknown chemicals or species.

\subsection{Metrics}
\label{sec:pap3_metrics}
We use two metrics to evaluate the performance of the models, coefficient of determination ($R^2$) and categorical accuracy (CA).
$R^2$ is defined as 
\begin{align}
    R^2 = 1 - \frac{\sum_i (y_i-\hat{y}_i)^2}{\sum_i (y_i-\overline{y})^2}
\end{align}
where $y_i$ is the true value, $\hat{y}_i$ is the predicted value, and $\overline{y}$ is the mean true value. Note that, this metric is calculated based on log-normalized concentrations.
This limits the effects of outliers. 

Categorical accuracy takes into account regularity requirements as defined in \citet{epa_classification} where categories \textit{very toxic} ($\leq 1mg/L$), \textit{toxic} ($> 1mg/L\ \wedge \leq 10mg/L$), \textit{harmful} ($> 10mg/L\ \wedge \leq 100mg/L$), and \textit{maybe harmful} ($>100mg/L$). We define categorical accuracy as 
\begin{align}
    CA = \frac{1}{N}\sum_i \frac{1}{1+|c_i-\hat{c}_i|},
\end{align}
where $c_i$ and $\hat{c}_i$ is the true and predicted categories, respectively. $N$ is the number of data points. This metric does not only value correct predictions but still gives, \eg $1/2$ to one category wrong prediction. Note that, due to this property the lower bound of the metric is $1/(1 + |1 - 4|)) = 1/4$ given the four categories defined above. 

\subsection{Gap-filling}
In this section, we present the results from gap-filling for missing chemical and species data.

\begin{figure}
     \centering
     \begin{subfigure}[b]{0.49\textwidth}
         \centering
         \includegraphics[width=\textwidth]{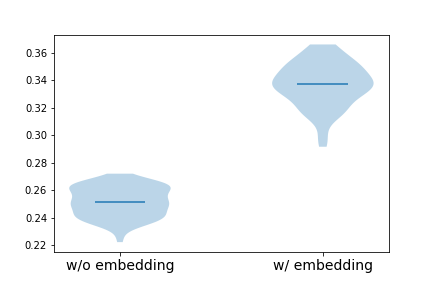}
         \caption{$R^2$}
         \label{fig:equal_prop_species_division_only_organic_predictions}
     \end{subfigure}
     \hfill
     \begin{subfigure}[b]{0.49\textwidth}
         \centering
         \includegraphics[width=\textwidth]{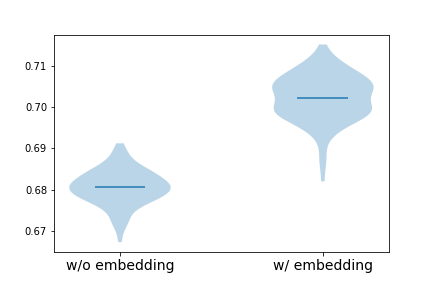}
         \caption{Categorical accuracy.}
         \label{fig:equal_prop_species_division_only_organic_predictions_cat_accuracy}
     \end{subfigure}
     \caption{Average and standard deviation of metrics in the species gap-filling prediction. \textit{w/o embedding} are results with random projection while \textit{w/ embedding} indicate the use of KG embeddings.}
     \label{fig:species_results}
\end{figure}

\begin{figure}
     \centering
     \begin{subfigure}[b]{0.49\textwidth}
         \centering
         \includegraphics[width=\textwidth]{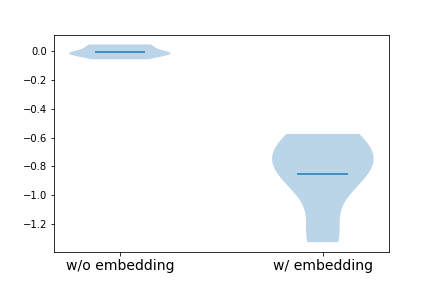}
         \caption{$R^2$}
         \label{fig:equal_prop_fp_sim_cluster_predictions_r2}
     \end{subfigure}
     \hfill
     \begin{subfigure}[b]{0.49\textwidth}
         \centering
         \includegraphics[width=\textwidth]{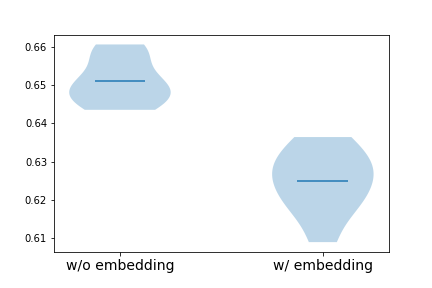}
         \caption{Categorical accuracy.}
         \label{fig:equal_prop_fp_sim_cluster_predictions_cat_accuracy}
     \end{subfigure}
     \caption{Average and standard deviation of $R^2$, in Figure \ref{fig:equal_prop_fp_sim_cluster_predictions_r2}, and Categorical accuracy ($CA$), in Figure \ref{fig:equal_prop_fp_sim_cluster_predictions_cat_accuracy}, in chemical gap-filling prediction. \textit{w/o embedding} are results with random projection while \textit{w/ embedding} indicate the use of KG embeddings.}
     \label{fig:chemical_results}
\end{figure}

\begin{figure}
    \centering
    \includegraphics[width=0.95\textwidth]{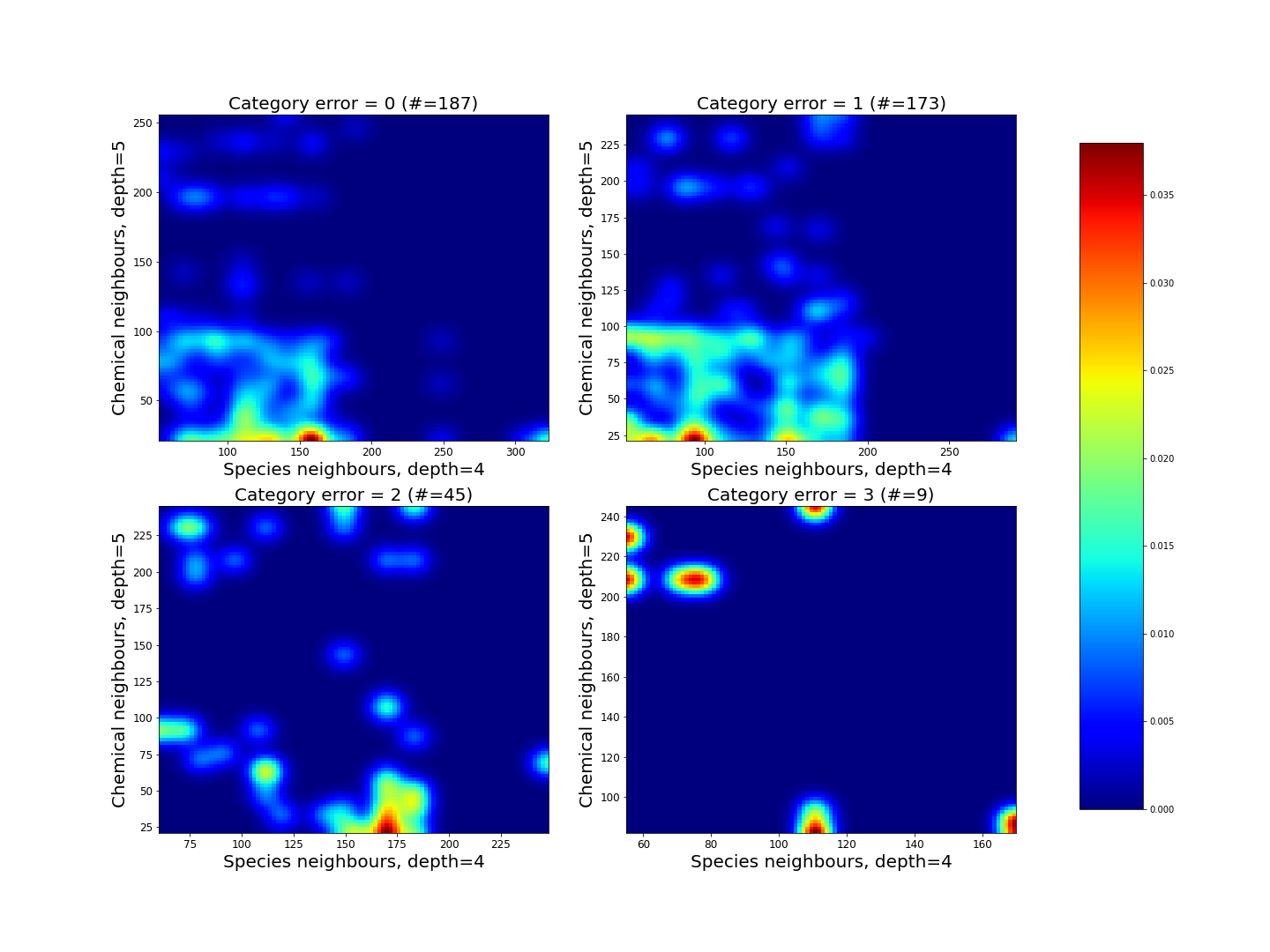}
    \caption{Density map of predictions (corresponds to Figure \ref{fig:equal_prop_species_division_only_organic_predictions}) as a function of number of neighbours (with effect data) within a graph depth of prediction chemical and species. The number of predictions in each plot is show as \eg $\# = 259$. The neighbourhood depth is shown as \eg $depth=9$. Note that, the axis are not necessary equal in all plots.}
    \label{fig:equal_prop_species_division_only_organic_symbolic_plot}
\end{figure}

Figure \ref{fig:species_results} show the results for species gap-filling while Figure \ref{fig:chemical_results} show the results for chemical gap-filling.  

Figure \ref{fig:equal_prop_species_division_only_organic_predictions} show an improvement of $\sim 0.09$ in $R^2$ using the knowledge graph embeddings. The categorical accuracy in Figure  \ref{fig:equal_prop_species_division_only_organic_predictions_cat_accuracy} is also improved, albeit, not as large change as for $R^2$. The relatively low standard deviation is down to the choice of the SVM as prediction model. 

We show the same plots for chemical gap-filling in Figure \ref{fig:chemical_results}. For this case, the prediction results without embeddings are equivalent to predicting prior mean, while embeddings introduce more noise which leads to negative $R^2$ value and $0.03$ lower $CA$. This indicates that the chemical gap-filling task is more difficult, and either the metadata in TERA does not describe chemicals adequately or the number of training samples needs to be increased, likely both. 


\subsection{Explanation}
In this section, we present several analysis using KGE to explain predictions. 

\begin{figure}
    \centering
    \includegraphics[width=0.95\textwidth]{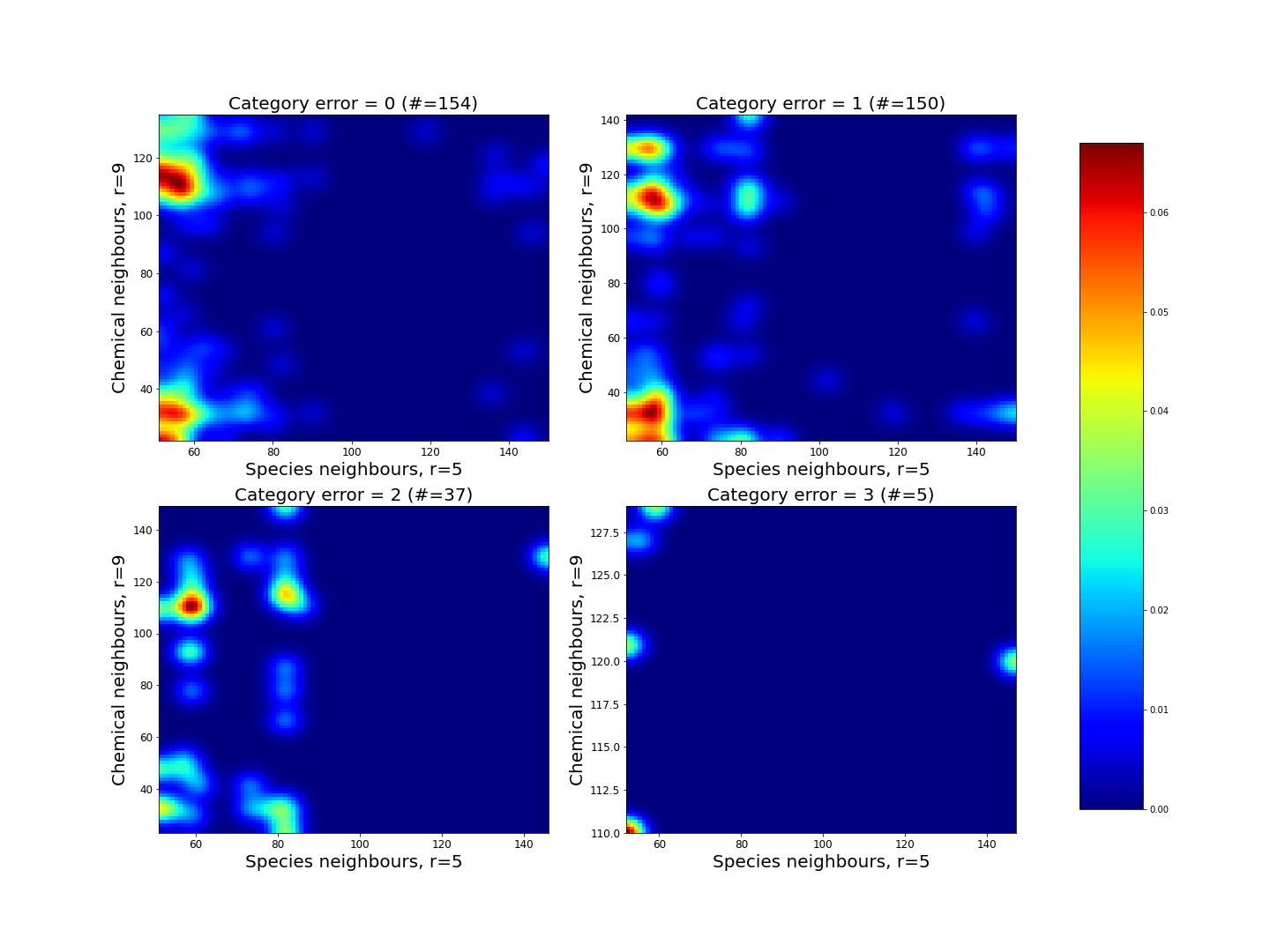}
    \caption{Density map of predictions (corresponds to Figure \ref{fig:equal_prop_species_division_only_organic_predictions}) as a function of number of neighbours within a radius to the prediction chemical-species pair. The number of predictions in each plot is show as \eg $\# = 259$. The neighbourhood radius is shown as \eg $r=9$. Note that, the axis are not necessary equal in all plots.}
    \label{fig:equal_prop_species_division_only_organic_temp_plot}
\end{figure}

Figure \ref{fig:equal_prop_species_division_only_organic_temp_plot} is generated by considering the categorical error of each individual prediction (\ie $0$ - correctly predicted, $1$ - one category error, $2$ - two category error or $3$ - three category error) and the number of neighbours the prediction pair (species and chemical) has closer than a certain radius.~\footnote{Note that, the majority of predictions have errors of $0$ or $1$ which makes the other figures less interpretable.} 
The categorical error is calculated as the 
difference between the true and the predicted category (\ie very toxic, toxic, harmful, and maybe harmful).
The density map is highly dependent on this radius and we have biased the radius to show interesting patterns. A larger radius would include more entities and \emph{smear} the density plots, while a smaller radius would include less entities and, therefore, densities would be \emph{patchy} and difficult to show. 
Figure \ref{fig:equal_prop_species_division_only_organic_symbolic_plot} show the same but defining neighbours as leaf nodes in the graph with data less than \emph{depth} (number of hierarchical levels to closest ancestor) away from prediction pair. This weights the data density more than the KG density in the calculation. Note that, the axis are not the same across plots.

Comparing category error $0$ and $1$ in Figure \ref{fig:equal_prop_species_division_only_organic_temp_plot} we can see (weakly) that the number of chemical neighbour density is similar, while the distribution of species distribution is wider where the error is fairly larger. This indicates that in the case of lack of species data, neighbourhood species density is the most important. 
This is not necessary surprising since a majority of the chemical side in a species-chemical is known and therefore, less emphasis is put on chemicals by the models. 

Figure \ref{fig:equal_prop_species_division_only_organic_symbolic_plot} shows similar patterns for all category errors, albeit a slightly more dense chemical neighbourhood for category error $1$. 
Figure \ref{fig:equal_prop_species_division_only_organic_symbolic_plot} also shows dispersion in both directions when moving from no to a small error (Categorical error 1). This can indicate the importance of data density for both chemical and species in this prediction task. The is expected as prediction methods used always perform better when more data is available. 
The larger errors show similar distributions, to each other, albeit at different scales (as the axis are different). This is a indication that certain structures are more important than others to good prediction performance.

\begin{figure}
    \centering
    \includegraphics[width=0.95\textwidth]{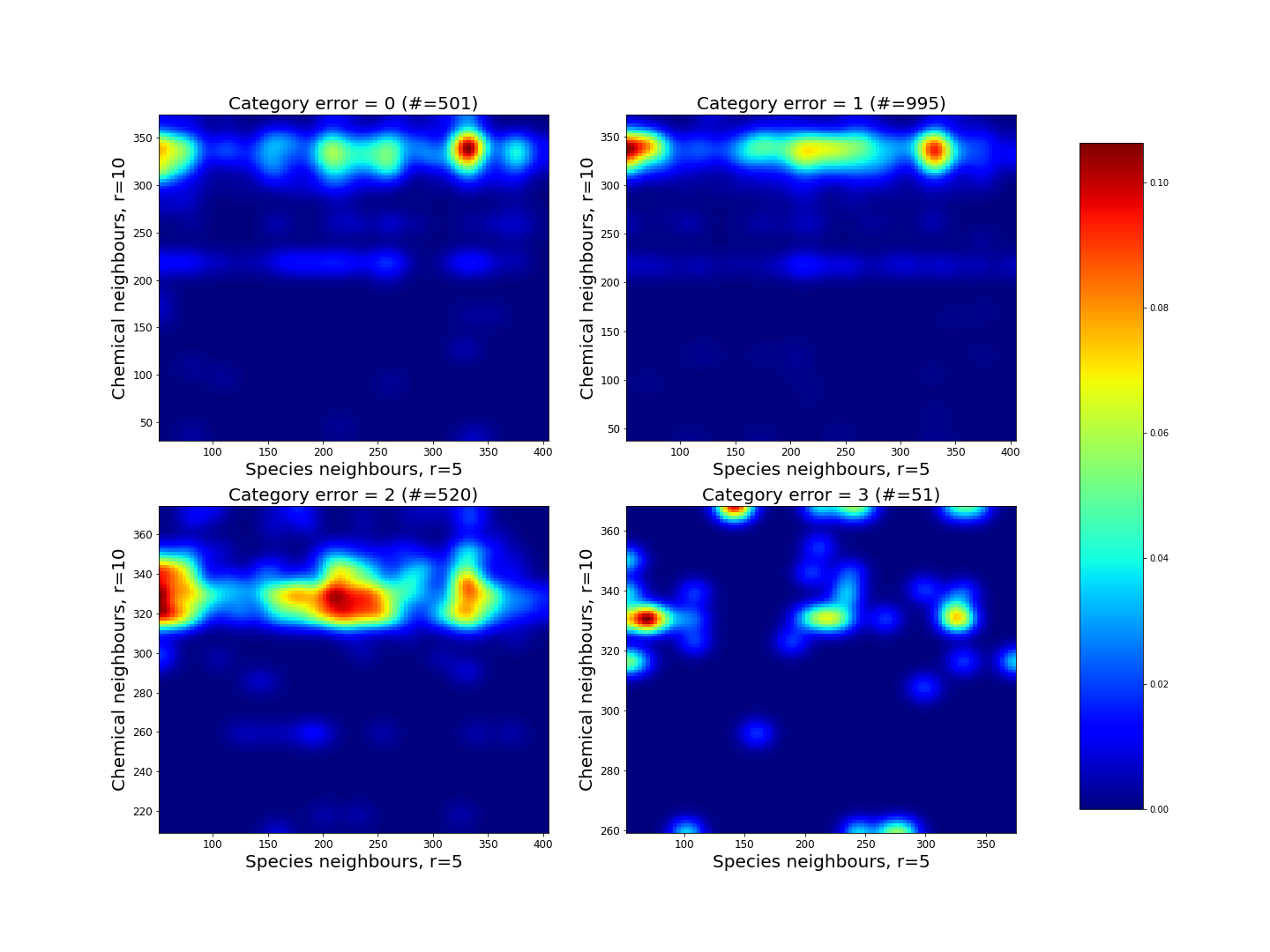}
    \caption{Density map of predictions (corresponds to Figure \ref{fig:chemical_results}) as a function of number of neighbours within a radius to the prediction chemical-species pair.  The number of predictions in each plot is show as \eg $\# = 501$. The neighbourhood radius is shown as \eg $r=10$. Note that, the axis are not necessary equal in all plots.}
    \label{fig:equal_prop_fp_sim_cluster_only_organic_temp_plot}
\end{figure}

Figure \ref{fig:equal_prop_fp_sim_cluster_only_organic_temp_plot} show the categorical error and the corresponding locations of each prediction in (chemical and species) neighbourhood space. 
Comparing bottom right of each panel, we see that there is a lower error where there is more diversity in neighbours, \ie number of neighbours is spread over a larger area as seen bottom left compared to top right in Figure \ref{fig:equal_prop_fp_sim_cluster_only_organic_temp_plot}. The difference is subtle, but is backed up by a small correlation between the number of chemical neighbours and mean absolute error ($0.08$, $p<0.05$). 

Another way to analyse the content of Figure \ref{fig:equal_prop_fp_sim_cluster_only_organic_temp_plot} is to train a model to predict error based on the neighbourhood (density). We take the vectors representing the distances to all other chemicals and species, and the absolute error of the pairs prediction and train a random forest model on this data. The model is able to predict the error with a mean error of $0.59\pm 0.04$ (over 100, 80/20 train/test random runs) which is less than the extremes of the standard deviation shown in Figure \ref{fig:std_input}. This error model can be used to generate confidence measures for the effect prediction model. 

\subsection{Common facts}
Common facts (triples) among data points can be a good way to analyse if a prediction is trustworthy. 
Table \ref{tab:common_facts} shows selected examples of common facts among the three closest neighbours to test chemical and species. We show two good, two moderate and one bad prediction. Table \ref{tab:common_facts} can be seen in conjunction with Table \ref{tab:corr_facts}, which indicates a (small) correlation between shared facts and the absolute error of a prediction. 
The correlation increase with the number of neighbours up to 11 neighbours and thereafter stagnates. This is down to the absence of shared facts at such distance away from the chemical-species pair. 

The correlation between number of facts and prediction error is positive for chemical data, \ie the more facts added the larger the error, indicating potential noise or not relevant facts for prediction in the current version of the chemical sub-KG of TERA. Moreover, the correlation of species data is negative, \ie more data\footnote{Not necessary more as this is common facts, but more concentrated data.} gives better performance which is expected. 
Table \ref{tab:common_facts} show certain aspects of this. \eg \texttt{Bluegill} in the first row shares both geo-region and habitat with other close entities which leads to good performance. While \texttt{Monkey River Prawns} present in Australia and P.R. China does not inspire confidence in the prediction as this does not define any descriptive features of the species, but rather superficial features of low importance for toxicity. For the moderate results we can see some noise (missing or wrong triples) in the KG, \eg \texttt{Fathead minnow} is present in \texttt{Mexico} (in TERA), however, in reality it is present in the greater Nearctic region (wrong in EOL and, therefore, also wrong in TERA).

Overall, we can see that there is little shared data among chemical neighbours which emphasises that the long hierarchical chains (which are not investigated here) of the graph might be more important.

\input{commonfacts}

\begin{table}[]
    \centering
    \begin{tabular}{c|c|c}
        $n$ & correlation & p-value \\\hline 
        $2$ & $0.08/0.10$ & $0.16/0.8$ \\\hline 
        $3$ & $0.05/0.05$ & $0.41/0.34$ \\\hline 
        $5$ & $0.12/0.04$ & $0.03/0.51$ \\\hline 
        $7$ & $0.11/(-)0.06$ & $0.05/0.32$ \\\hline 
        $9$ & $0.12/(-)0.14$ & $0.03/0.01$ \\\hline
        $11$ & $0.17/(-)0.14$ & $0.00/0.01$ \\
    \end{tabular}
    \caption{Correlation between number of facts (Table \ref{tab:common_facts}) with $n$ neighbours and absolute error. The two numbers represent the correlation and p-value for experimental chemical and species, respectively.}
    \label{tab:corr_facts}
\end{table}

\section{Discussion}
\label{sec:pap3_discussion}

We have presented results that show the benefit of using knowledge graph embeddings in predicting the correct toxicity of missing data within similar chemical-species groups (\ie read-across). 
In addition, we have presented several methods for providing quantitative and qualitative insight of these predictions. 

The results from this study are encouraging when applied in this example case. We improve results with around 40\% in $R^2$ ($0.25 \to 0.34$) the setting of gap-filling unknown species with organic chemicals. Along with the results, three explanations where provided. The first provide an overall view of graph density in relation to the prediction performance. This can be used to assess the quality of the knowledge graph, the embeddings, and the trustworthiness of the adverse effect data itself. 

The error prediction can directly be used to assess the confidence of the individual predictions. This can further provide species or chemical groups where the data density can be improved, \ie predicted error can be used as a ranking of which laboratory experiments will be overall beneficial for the effect prediction model. 

Comparing results of gap-filling unknown species vs. gap-filling unknown chemicals we can deduce that prediction of unknown chemicals is a harder task as classifications and functional groups do not describe chemicals as with the same resolution as hierarchy and traits define species. 
This is where and investigation into which parts of TERA contributes and counteract the predictions is needed. 
Other information such as the chemical molecules added as a OWL based sub-graphs (\eg \citet{Hastings2010-06}) can also be added to TERA to improve the granularity of the KG. 

This lack of complete data is the biggest caveat in this work. Using automated knowledge extraction and integration gathers available data, but does not necessary take the specific species or chemical domains into account. This could be solved with a bottom-up approach seeing where the baseline model falls short and adding data to the knowledge graph accordingly. 

Given the discrepancy in the training data as shown in Figure \ref{fig:std_input}, we are satisfied with the performance of the models in for initial screening purposes. Consistent effect data is key for this method to outperform narrow domain models across multiple application domains, however, these datasets are not available currently and beyond the scope of this study.

\section{Conclusion}
\label{sec:pap3_conclusion}
In this work we have set up a framework for using knowledge graphs in ecotoxicological effect prediction, especially as a initial screening tool for unknown and rare chemicals or species. 

This framework includes the use of tools to integrate data sources into a domain specific knowledge graph we call TERA. The knowledge graph entities can be represented in a vector space using knowledge graph embedding models. This enables the use of chemicals and species in TERA directly in machine learning models. Moreover, the prediction results can be improved by using embeddings over a naive baseline and is the basis of explanations of the predictions. However, we found that chemical hierarchies are not as descriptive as species hierarchies when applied in this prediction task. 

\subsection{Future work}
Previous \citep{Myklebust2019KnowledgeGE, swj_paper} and present studies have shown one way knowledge graphs can be included in effect prediction. And we show how a KG can help gain insight into the predictions. However, this ground work is just the beginning of integrating these methodologies into hazard assessment approaches. 
The next steps is to get the tools and prediction models integrated into computational pipelines to support effect predictions and read-across. This would potentially reduce the application domain further which may require additional effect data sources to be integrated into TERA. Furthermore, more effect data will be a benefit for a more extensive knowledge graph and we plan to integrate yet more resources into TERA. On the other hand, extension of KGE models to handle hierarchies are expected to increase performance. 

This use-case is also well suited to test improvements of new semantic web tools, \eg ontology alignment, due to the size of TERA; and KGE models, due to the unique structure of TERA. 

\subsection{Resources}
The resources for this paper are available at
\url{https://github.com/NIVA-Knowledge-Graph/kge_ecotox_regression}.

The source code for TERA is available at \url{https://github.com/NIVA-Knowledge-Graph/TERA}.

\medskip
\noindent
\textbf{Acknowledgements}.
This work is supported by the grant 272414 from the Research Council of Norway (RCN), the MixRisk project (Research Council of Norway, project 268294), SIRIUS Centre for Scalable Data Access (Research Council of Norway, project 237889), Samsung Research UK, Siemens AG, and the EPSRC projects AnaLOG (EP/P025943/1), OASIS (EP/S032347/1), UK FIRES (EP/S019111/1), the AIDA project (Alan Turing Institute), by NIVAs Computational Toxicology Program, NCTP (\url{niva.no/nctp}), and donated weekends.

\bibliographystyle{plainnat}
\bibliography{bibliography}
 





\appendix
\section{Method Details}
\label{sec:method_appendix}
This appendix provides details for Section \ref{sec:pap3_method}. 

\subsection{Effect data query}
\label{sec:query_appendix}

\begin{lstlisting}[language=sparql, caption=Querying experimental data with SPARQL \label{lst:query}]
SELECT ?species ?chem ?conc ?unit WHERE {
        ?test ns:hasResult [ns:endpoint ?end ;
                            ns:effect effect:MOR ;
                            ns:concentration [
                                rdf:value ?conc ;
                                unit:units ?unit 
                            ]] .
        ?test ns:species ?species .
        ?test ns:chemical ?chem .
        ?test ns:studyDuration [rdf:value ?sd ;
                                unit:units ?cu] ] .
    FILTER (?unit in (unit:MicrogramPerLitre, 
                        unit:MilligramPerLitre))
    FILTER (?end in (endpoint:LC50, 
                        endpoint:LD50, 
                        endpoint:EC50) )
    FILTER (?cu = unit:Hour)
    FILTER (?sd >= 24 || ?sd <= 96)
    }
\end{lstlisting}

Query \ref{lst:query} looks for tests (\ie experiments) in ECOTOX with results which has mortality as effect and a endpoint in $\{LC_{50},LD_{50},EC_{50}\}$. \footnote{We consider $LC_{50} \land MOR \equiv LD_{50} \land MOR \equiv EC_{50} \land MOR$.} The query also filters on the units used for the chemical concentration, this is a \textit{cheap} trick to filter on only results where the chemical is added to the aquatic environment and not given orally. Finally, we filter test with a duration between 24 and 96 hours. 

\subsection{Optimizing ComplEx}
\label{sec:complex_appendix}

The KGE model is optimized by minimizing the pointwize logistic loss 
\begin{align}
    L = \sum_{t\in KG} \func{log}(1+\func{exp}(-y_{\vec{t}} S(\vec{t}))
\end{align} 
where $y_{\vec{t}} \in \{-1,1\}$ is the label of the triple, $1$ for true, $-1$ for false. The false triples are created by replacing the subject and/or object of triples in the KG with random entities from the KG.

\begin{figure}
    \centering
    \includegraphics[width=0.95\textwidth]{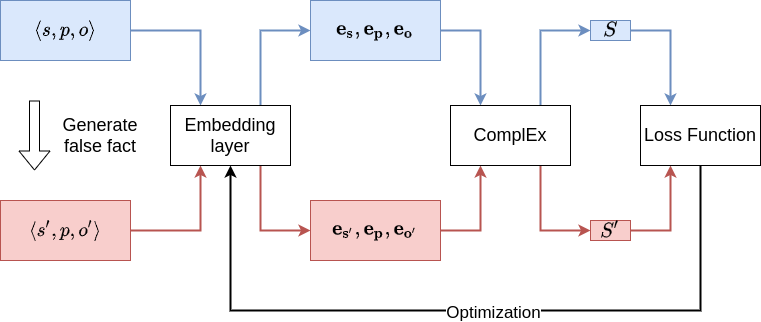}
    \caption{Optimization of ComplEx. Each triple $t$ in TERA is corrupted to $t^\prime$; vector representations (random at start) are extracted for the triples; the score for $t$ and $t^\prime$ is calculated; and the loss function is calculated; finally, updated to the embedding layer is calculated and applied.}
    \label{fig:OptimizingKGE}
\end{figure}

Figure \ref{fig:OptimizingKGE} show the stepwise optimization of ComplEx. A false fact is generated randomly from the true fact by permuting subject and/or object. Thereafter, the true fact follow the blue colour while the false fact follows the red colour. The two scores are evaluated in the loss function and a correction of the embedding layer is calculated and applied by the optimizer, we use the Adam optimizer \citep{Adam_article}.

\subsection{Support Vector Machines}
\label{sec:svm_appendix}

A SVM is normally used in classification problems where the goal is to maximize the distance between the decision boundary (class divide) and the individual points. Regression can be performed using a SVM by considering the margin from the decision boundary as a real number and not a strict boundary. 

SVMs are strictly linear, however, applying a kernel can be done to make it non-linear. This projects the data points into a higher dimensional space where the problem becomes linear. In this work, we use a radial basis function kernel defined as 
\begin{align*}
    K(\vec{x},\vec{x^\prime}) = \exp(-\gamma ||\vec{x}-\vec{x^\prime}||^2)
\end{align*}
where $\gamma = 1/2\sigma$, and $\vec{x}$ and $\vec{x^\prime}$ are two different datapoints. 

\subsection{Explanation Algorithms}
\label{sec:algo_appendix}

In this section, we will describe the algorithms in plain text. The algorithms are available in the projects GitHub repository. \footnote{\url{https://github.com/NIVA-Knowledge-Graph/kge_ecotox_regression}}

\textit{Neighbourhood density.}
Based on the embeddings we calculate the distance from prediction chemical and species to other entities in the KG as in \eqref{eq:sim}. Thereafter, we find how many chemical entities are within $r_1$ of test chemical and species entities are within $r_2$ of test species. Figures \ref{fig:equal_prop_species_division_only_organic_temp_plot} and \ref{fig:equal_prop_fp_sim_cluster_only_organic_temp_plot} are produces by plotting the density of the number of neighbours in each error category, \ie 0,1,2,3 error. 

\textit{Error prediction.}
Based on the embeddings a similarity measure can be calculated. We define a similarity matrix as 
\begin{align}
    S_{i,j} = ||\vec{e}_i-\vec{e}_j||_2, \label{eq:sim}
\end{align}
\ie the Euclidean distance between the entity embeddings. 

From the prediction model we can gather an (mean absolute) error for each prediction. We train a random forest model (RF) to predict the error where the input is the similarity vectors from above. We use the same 5-fold validation as described in Section \ref{sec:pap3_method}.

\textit{Common facts.}
Based on \eqref{eq:sim} we can find the $n$ closest entities to an prediction entity (chemical or species). This set of $n$ closest entities and entity we call M. Then the common facts are defined as
\begin{align}
    CF = \bigcap_{e \in M} \{ \langle s,p,o \rangle | s\equiv e  \land \langle s,p,o \rangle \in KG\},
\end{align}
\ie the intersection of facts associated with each entity in $M$.

\end{document}

%% file: commonfacts.tex
\setlength{\tabcolsep}{2.0pt}
\begin{sidewaystable}[]
\caption{Table of selected predictions and their common facts to other data (3 closest neighbours in the KG). The columns shown the \textit{Error} of the prediction where absolute error $AE = log_{10}(\hat{y}) - log_{10}(y)$ ($y$ is concentration), categorical accuracy is defined in main text; \textit{Entity}, \textit{Predicate} and \textit{Entity/Literal} are the components of the shared triples. \textit{Entity/Literal} can contain multiple values if there are multiple objects for the particular subject-predicate pair.}
    \label{tab:common_facts}
    \begin{tabular}{c|c|c|c}
        \textit{Error} & \textit{Entity} & \textit{Predicate} & \textit{Entity/Literals} \\\hline 
        $AE=0.0026$ & \texttt{Bluegill} & \texttt{eol:Present} & \texttt{North Atlantic Ocean}, \texttt{North Pacific Ocean} \\
        $CA=1$ & & \texttt{eol:Habitat} & \texttt{freshwater biome} \\
        & \texttt{DDT}\footnote{1,1'-(2,2,2-Trichloroethylidene)bis(4-chlorobenzene)} & \texttt{ex:compoundIsHeavy} & \texttt{False} \\\hline
        $AE=0.0211$ & \texttt{Colorado Squawfish} & \texttt{eol:Habitat} & \texttt{freshwater biome}, \texttt{freshwater environment} \\
        $CA=1$ & & \texttt{rdf:type} & \texttt{Vertebrate} \\
        & \texttt{Permethrin} & \texttt{ex:compoundIsHeavy} & \texttt{False} \\\hline
        $AE=0.55$ & \texttt{Fathead Minnow} & \texttt{eol:Habitat} & \texttt{freshwater biome}, \texttt{large river biome}, \texttt{small river biome}  \\
        $CA=0.5$ & & \texttt{eol:Present} & \texttt{Mexico} \\
        & & \texttt{rdf:type} & \texttt{Vertebrate} \\
        & \texttt{Hydroquinone} & \texttt{ex:compoundIsHeavy} & \texttt{False} \\\hline
        $AE=2.73$ & \texttt{Western Toad} & \texttt{eol:Habitat} & \texttt{freshwater biome}, \texttt{woodland}, \texttt{pond} \\
        $CA=0.5$ & & \texttt{eol:Present} & \texttt{the United States of America} \\
        & & \texttt{obo:GAZ\_00000071}\footnote{Biogeographic realm} & \texttt{Nearctic realm} \\
        & & \texttt{rdf:type} & \texttt{Vertebrate} \\
        & \texttt{Permethrin} & \texttt{ex:compoundIsHeavy} & \texttt{False} \\\hline
        $AE=4.87$ & \texttt{Monkey River Prawn} & \texttt{eol:Present} & \texttt{Australia}, \texttt{P.R. China}, \\
        $CA=0.25$ & & \texttt{rdf:type} & \texttt{Invertebrate} \\
        & \texttt{Cypermethrin} & \texttt{ex:compoundIsHeavy} & \texttt{False} \\\hline
    \end{tabular}
\end{sidewaystable}